%% file: main.tex
\pdfoutput=1
\documentclass[10pt]{article}
\usepackage[letterpaper,margin=1in,headsep=14pt]{geometry}
\usepackage[T1]{fontenc}
\usepackage[utf8]{inputenc}
\usepackage{amsmath,amsthm,mathtools}
\usepackage[helvratio=0.92]{newtxtext}
\usepackage{newtxmath}
\usepackage{microtype}

\usepackage{booktabs,multirow,array}
\usepackage{graphicx}
\usepackage[table]{xcolor}
\usepackage[authoryear,round]{natbib}
\usepackage{algorithm}
\usepackage[noend]{algpseudocode}
\usepackage{enumitem}
\usepackage{placeins}
\usepackage{pifont}

\definecolor{ptsnavy}{HTML}{1F3A68}
\definecolor{ptsaccent}{HTML}{D9472B}
\definecolor{ptsgray}{HTML}{F3F4F6}
\definecolor{ptsrow}{HTML}{FDEEE8}
\definecolor{ptslink}{HTML}{24539A}
\usepackage[most]{tcolorbox}
\usepackage{titlesec}
\titleformat{\section}{\sffamily\bfseries\large\color{ptsnavy}}{\thesection}{0.6em}{}
\titleformat{\subsection}{\sffamily\bfseries\normalsize\color{ptsnavy}}{\thesubsection}{0.6em}{}
\titleformat{\paragraph}[runin]{\sffamily\bfseries\small}{}{}{}[.]
\titlespacing*{\section}{0pt}{10pt plus 3pt minus 2pt}{5pt plus 1pt}
\titlespacing*{\subsection}{0pt}{8pt plus 2pt minus 2pt}{3pt plus 1pt}
\titlespacing*{\paragraph}{0pt}{5pt plus 1pt minus 1pt}{0.6em}
\usepackage{caption}
\DeclareCaptionLabelSeparator{bar}{\enspace\textcolor{gray}{|}\enspace}
\usepackage{fancyhdr}
\fancypagestyle{first}{\fancyhf{}\fancyfoot[C]{\sffamily\footnotesize\thepage}}
\setlist{leftmargin=*,itemsep=1.5pt,topsep=2pt}
\usepackage[colorlinks=true,linkcolor=ptslink,citecolor=ptslink,urlcolor=ptslink]{hyperref}
\usepackage[capitalise,noabbrev]{cleveref}

\input{macros}
\newcommand{\cmark}{\textcolor{ptsaccent!0!green!55!black}{\ding{51}}}
\newcommand{\xmark}{\textcolor{red!70!black}{\ding{55}}}
\newcommand{\ours}{\rowcolor{ptsrow}}

\begin{document}
\thispagestyle{first}
\begin{center}
{\LARGE\bfseries Same Path, Less Time: Training-Free Fast Execution of\\[2pt] Vision-Language-Action Models under Inference Latency\par}
\vspace{14pt}
{\large Fuyi Chen\par}
\vspace{2pt}
{University of Nottingham Ningbo China\par}
\end{center}
\vspace{4pt}
\begin{center}{\large\bfseries Abstract}\end{center}
\vspace{-4pt}
\begin{quote}
\input{sections/abstract}
\end{quote}

\vspace{2pt}
\begin{figure}[h]
\centering
\includegraphics[width=\linewidth]{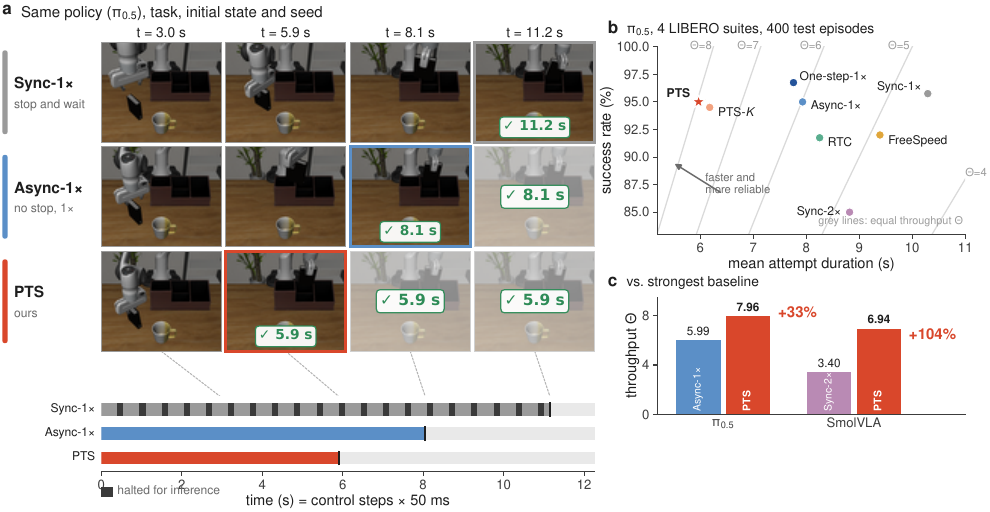}
\caption{\textbf{Same policy, same episode, less time.} \textbf{a},~Frozen \pifive{} on a LIBERO-Long task (same initial state and seed; columns at equal times, counted in 50\,ms control steps; the time line marks steps halted for inference). Stop-and-wait (Sync-1$\times$) halts at every inference, asynchronous execution (Async-1$\times$) never halts but stays at demonstration speed, and \pts{} never halts and traverses the planned path faster. \textbf{b},~Success rate versus mean attempt duration on the full \pifive{} test set; gray lines have equal throughput $\Theta=1000\,P/\E[T]$. \textbf{c},~Pooled throughput of \pts{} against the strongest deployable baseline of each model.}
\label{fig:teaser}
\end{figure}

\input{sections/intro}
\input{sections/related}
\input{sections/setup}
\input{sections/method}
\input{sections/experiments}
\input{sections/discussion}

\FloatBarrier
\bibliographystyle{plainnat}
\bibliography{refs}

\clearpage
\appendix
\input{appendix/analysis}
\input{appendix/implementation}
\input{appendix/results}

\end{document}

%% file: macros.tex
\newtheorem{proposition}{Proposition}

\newcommand{\pts}{PTS}
\newcommand{\ptsk}{PTS-$K$}
\newcommand{\pizero}{$\pi_0$}
\newcommand{\pifive}{$\pi_{0.5}$}
\newcommand{\smol}{SmolVLA}
\newcommand{\Rl}[1]{\textbf{R#1}}

\newcommand{\E}{\mathbb{E}}

\newcommand{\Rbody}{R_{\mathrm{body}}}
\newcommand{\Rfeas}{R_{\mathrm{feas}}}
\newcommand{\rbar}{\bar r}
\newcommand{\rbody}{r_{\mathrm{body}}}
\newcommand{\TVL}{T_{\mathrm{VL}}}
\newcommand{\Tstep}{T_{\mathrm{step}}}
\newcommand{\Pstar}{P^{\ast}}
\newcommand{\tr}{\operatorname{tr}}
\newcommand{\Cov}{\operatorname{Cov}}
\newcommand{\ci}[2]{[#1,\,#2]}

%% file: sections/abstract.tex
Vision-language-action (VLA) models predict chunks of future actions, yet deployed robots waste time in two ways that are independent of model capability: they halt while the next chunk is inferred, and they move at the speed of slow, cautious demonstrations. Existing remedies remove one loss each, and stacking them fails, because asynchrony and speed-up each change the correspondence between planned progress and executed motion. We present \emph{Path-preserving Time Scaling} (\pts{}), a training-free execution layer for frozen action-chunking policies built on one principle: \emph{the robot's motion is a time reparameterization of a reference path, and the path changes only in response to new observations.} A path layer decodes every replan deterministically in one step and enters it at the progress consumed during inference. A time layer sums the planned increments, keeps the arm at 1$\times$ while the gripper moves, and caps the rate by the actuator's linear range and by chunk feasibility under latency. We show that misaligned entry and stochastic sampling add a bias and a variance to the continuation error, and derive a break-even success rate that states when speeding up pays off. On LIBERO with two frozen VLAs under injected inference latency, \pts{} raises task throughput over the strongest deployable baseline by 33\% for \pifive{} and 104\% for \smol{} while matching or improving the success rate, is best on all six model--suite combinations, and exceeds zero-latency 1$\times$ execution. Against asynchronous 1$\times$ execution with the same one-step decoder, the time layer alone adds 28\% and 38\%. On LIBERO-Dynamic it improves on stop-and-wait execution by 38\% and is comparable to asynchronous 1$\times$ execution.

%% file: sections/intro.tex
\section{Introduction}
\label{sec:intro}

Vision-language-action models (VLAs) \citep{brohan2023rt2,kim2024openvla,black2024pi0,pi2025pi05,shukor2025smolvla} predict \emph{action chunks}, short sequences of future actions \citep{zhao2023act,chi2023diffusion}. The default deployment is synchronous: infer a chunk, execute part of it, stop, and infer again. This wastes time in two ways that do not depend on how capable the model is. First, the robot stands still during every inference: if one \pifive{} inference takes about 150\,ms, a 20\,Hz robot pauses for three steps after every eight it executes. Second, execution is locked to the speed of the demonstrations, which are typically slow, cautious teleoperation. \Cref{fig:teaser}a shows both losses in one episode. This paper removes both for a \emph{frozen} policy, without any training.

\begin{figure}[t]
\centering
\includegraphics[width=\linewidth]{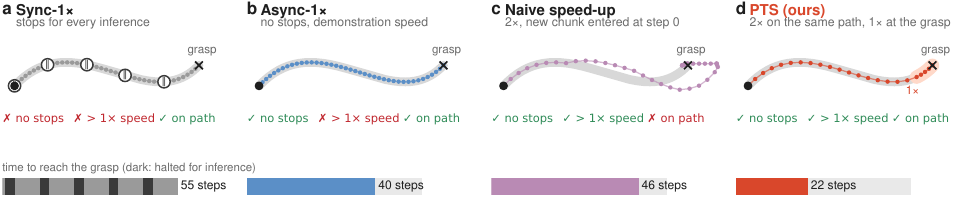}
\caption{\textbf{Four ways to execute a chunking policy under inference latency} (2-D schematic; dots are positions at successive control steps, so spacing is speed). \textbf{a},~Stop-and-wait halts at every inference (rings). \textbf{b},~Asynchronous 1$\times$ never halts but moves at demonstration speed. \textbf{c},~Naive asynchronous speed-up enters each chunk at its first step, replays motion already executed, and overshoots. \textbf{d},~\pts{} enters at the consumed progress, traverses the same path at twice the rate, and returns to 1$\times$ around the gripper event (shaded).}
\label{fig:concept}
\end{figure}

\paragraph{Why the two fixes do not compose}
Each loss has a known remedy. Asynchronous execution, e.g., real-time chunking (RTC) \citep{black2025rtc}, keeps the robot moving during inference but executes at 1$\times$. Training-free speed-up such as FreeSpeed \citep{hu2026freespeed} amplifies the actions but assumes synchronous execution. Stacking the two fails: run asynchronously, FreeSpeed drops the success rate of \pifive{} on our LIBERO-Spatial development set from 45/50 to 20/50, even with the same gripper handling as our method; keeping its speed profile but summing the planned increments instead of resampling them restores 47/50. The cause is bookkeeping. A chunk is a plan that starts at the moment its image was taken, and by the time it arrives the robot has already consumed part of it; speed-up changes how much. If the executor loses track of how much of the plan has actually been executed, the robot repeats or skips motion and leaves the planned path (\cref{fig:concept}c).

\paragraph{Key idea: separate the path from its timing}
An action chunk defines a planned path. If execution changes only the \emph{timing} along that path, i.e., the robot follows the path the model planned but traverses it faster and never stops during inference, then speeding up reduces to shortening time inside a feasible set. This separation of path and time underlies time-optimal path parameterization (TOPP) in robotics \citep{bobrow1985time,shin1985minimum,pham2018toppra}. Learned policies add two ingredients that classical TOPP lacks: closed-loop \emph{replanning} makes the path a concatenation of chunks, and \emph{inference latency} means the robot has moved on by the time a new chunk arrives. Execution therefore has to answer two questions. \emph{Which path} should the robot follow at each instant? Our answer is that the path should change only because of new information: decode the plan for the newest observation deterministically, and enter it at the progress actually consumed during inference. \emph{How fast} should it traverse that path? As fast as the actuator's linear range and the chunk's lifetime allow, but at 1$\times$ while the gripper moves, with commands formed by summing the planned increments.

\paragraph{Results}
We call the resulting method \emph{Path-preserving Time Scaling} (\pts{}, \cref{fig:overview}). On the four LIBERO suites with frozen \pifive{} and \smol{} under injected inference latency, \pts{} raises the pooled task throughput by 33\% and 104\% over the strongest deployable baseline, matching the success rate on \pifive{} and raising it on \smol{}, and is best on all six model--suite combinations (\cref{fig:teaser}b,c). Both layers contribute: with the same one-step decoder \citep{chen2026simple}, the time layer adds 28\% and 38\% over asynchronous 1$\times$ execution, and deterministic decoding raises \smol{}'s success rate from 0.53 to 0.805 under the same timing rules.

\paragraph{Contributions}
\begin{itemize}
\item \textbf{Method.} \pts{}, a training-free execution layer for frozen VLAs that output relative-increment chunks. It consists of one path-layer rule (rolling continuation with deterministic one-step decoding) and three time-layer rules (incremental summation, gripper synchronization, rate limit), and has one tuned hyperparameter, the rate cap $\rbar$. For flow-matching policies it also makes inference cheaper than $K$-step sampling.
\item \textbf{Analysis.} A decomposition of the continuation error showing that misaligned entry adds a bias and stochastic sampling a variance (\cref{prop:decomp}), and a break-even success rate that states when speeding up pays off (\cref{prop:throughput}).
\item \textbf{Evaluation.} Two VLAs, four LIBERO suites, six baselines, layer and rule ablations, and a dynamic benchmark (6{,}160 test episodes), plus studies of rate selection, actuator strength, and latency.
\end{itemize}

%% file: sections/related.tex
\section{Related Work}
\label{sec:related}

\paragraph{Inference latency and asynchronous execution}
Predicting chunks reduces compounding error \citep{zhao2023act,chi2023diffusion}, and most VLAs adopt this output, generated by flow matching \citep{black2024pi0,pi2025pi05,shukor2025smolvla} or decoded from action tokens \citep{pertsch2025fast}. Asynchronous execution keeps executing the old chunk during inference, at the price of seams between chunks and of plans conditioned on outdated observations. Training-free methods make the new chunk agree with the old one: BID selects the most consistent of several samples \citep{liu2025bid}, RTC freezes the executed prefix by guided inpainting \citep{black2025rtc}, FreeBar replaces guidance by a closed-form barrier \citep{yu2026freebar}, PACT-VLA verifies candidates speculatively \citep{zhu2026pact}, and PICES adapts the number of executed steps \citep{lee2026pices}; \smol{} ships a queue-based asynchronous stack \citep{shukor2025smolvla}. Training-based methods condition on the executed prefix \citep{black2025ttrtc}, on future states \citep{tang2025vlash}, or on the newest observation through a correction head \citep{sendai2025a2c2}, or add adapters, preference learning, or streaming decoding \citep{jiang2026futurertc,zhu2026deflect,lu2026faster,li2026flashvla,yang2026eaglevla}; \citet{agouzoul2026async} compares them under one protocol. Apart from VLASH, which also speeds up execution (discussed below), these methods execute at 1$\times$. The path layer of \pts{} approaches inter-chunk consistency differently: instead of steering a random sample toward the old chunk, it removes sampling randomness that carries no new information and enters the new plan at the consumed progress. It needs no guidance (RTC guidance multiplies inference time by about 1.8 in our implementation), and one-step decoding makes inference faster.

\paragraph{Faster-than-demonstration execution}
One route retrains the policy on accelerated or downsampled demonstrations \citep{guo2025demospeedup,kim2025espada,cao2026adatempo}, with speed as an input \citep{jing2026tempovla,nam2025speedaug,jeon2026pacevla}, or with a learned scheduler \citep{wu2026speedup,yang2026realtimevla2}. A second changes the imitation target: SAIL imitates reached rather than commanded states and requires a chunk to outlast the inference time \citep{arachchige2025sail}, and VOLT segments trajectories to decide what may be accelerated \citep{ramirez2026volt}. A third adjusts speed at test time: FreeSpeed resamples the chunk at the target rate in synchronous execution and shrinks the step size with the directional inconsistency of adjacent increments \citep{hu2026freespeed}, RoboPace retimes chunks with TOPP-RA under a learned contact speed limit \citep{shirasaka2026robopace}, and VLASH, a training-based method, additionally speeds up execution via action quantization \citep{tang2025vlash}. We are not aware of prior work that handles speed-up and inference latency jointly for a frozen VLA without training. Unlike FreeSpeed, \pts{} sums increments instead of resampling, does not speed up while the gripper moves, and executes asynchronously with entry at the consumed progress.

\paragraph{One-step action generation}
Consistency and shortcut models train few-step samplers \citep{song2023consistency,frans2025shortcut}; for robot policies, consistency distillation \citep{prasad2024consistency} and Mean-Flow \citep{chen2026meanflow} yield one-step VLAs. \citet{pan2025mip} argue that generative control does not owe its advantage mainly to multimodality. \citet{chen2026simple} show that one Euler step from the noise end returns the conditional mean for an ideal velocity field and evaluate off-the-shelf VLAs with one step under synchronous execution. We adopt this decoder in the path layer (\cref{sec:path}); our contribution is its coordination with asynchronous execution and time scaling.

\paragraph{Path parameterization and receding-horizon control}
TOPP computes the minimum-time traversal of a fixed path under dynamic constraints \citep{bobrow1985time,shin1985minimum}, by convex optimization \citep{verscheure2009time}, numerical integration \citep{kunz2012time}, or reachability analysis \citep{pham2018toppra}; under pointwise decoupled bounds the optimum saturates the bound. \pts{}'s time layer is this case applied to chunk execution (\cref{app:time}); what is new is where the bounds come from: the actuator's linear range, chunk feasibility under latency, and a rate cap selected from data. Receding-horizon control executes the beginning of each newly solved plan \citep{mayne2000constrained}; the path layer extends this practice to delayed inference and accelerated consumption of the plan.

%% file: sections/setup.tex
\section{Problem Formulation}
\label{sec:setup}

\paragraph{Policy and planned path}
The control period is $\Delta t=50$\,ms. Given an observation $o$, a frozen policy defines a distribution $\pi(\cdot\mid o)$ over chunks $A=(a_0,\dots,a_{H-1})$, where each action $a_i=(a_i^x,a_i^g)$ is a relative end-effector increment $a_i^x\in[-1,1]^6$ and a gripper command $a_i^g$; we write $\mu(o):=\E[A\mid o]$. We measure progress $\sigma\in[0,H)$ in plan steps. The planned velocity is $\alpha(\sigma):=a^x_{\lfloor\sigma\rfloor}$ and the cumulative planned command is $\Phi(\sigma):=\int_0^\sigma\alpha$. Under 1$\times$ execution, $a_t$ is executed at step $t$.

\paragraph{Flow-matching policies and latency}
\pizero{}, \pifive{}, and \smol{} generate chunks by flow matching \citep{lipman2023flow,liu2023rectified}: with $\varepsilon\sim\mathcal N(0,I)$ and $x_\tau=\tau\varepsilon+(1-\tau)A$, they learn $v_\theta(x,\tau\mid o)\approx\E[\varepsilon-A\mid x_\tau=x,o]$ and integrate from $\tau=1$ to $0$ with $K$ Euler steps (default $K=10$). One inference takes $\TVL+K\Tstep$ (the vision-language backbone once, the action expert $K$ times), i.e., $d_K:=\lceil(\TVL+K\Tstep)/\Delta t\rceil$ control steps.

\paragraph{Execution schedule}
An executor chooses a progress sequence $p_t$ with rate $\rho_t:=p_{t+1}-p_t\ge0$ and a command rule $u_t=\mathcal E(A,p_t,\rho_t)$. A new inference is requested every $s$ consumed plan steps; its result arrives $d$ control steps later, and in asynchronous execution the robot keeps moving in between. The $k$-th chunk defines a planned path $\Phi_k$, and the \emph{reference path} the robot follows is the concatenation of $\Phi_1,\Phi_2,\dots$ at the switching points. The executor therefore decides two things: which chunk and entry point define the reference path (the \emph{path}), and how fast it is consumed and converted to commands (the \emph{timing}).

\paragraph{Objective: task throughput}
Attempts are drawn independently from the distribution of tasks and initial states. We maximize the task throughput
\begin{equation}\label{eq:theta}
\Theta:=1000\cdot\frac{\E[Y]}{\E[T]},
\end{equation}
where $Y\in\{0,1\}$ indicates success and $T$ is the number of control steps of an attempt, including steps spent waiting for inference; $1000/\Theta$ is the average number of steps per success. $\Theta$ rewards speed and success jointly: faster execution that fails more often is not necessarily better (\cref{fig:teaser}b). It is a ratio of pooled means, not a mean of per-task ratios.

%% file: sections/method.tex
\section{Path-Preserving Time Scaling}
\label{sec:method}

\begin{figure}[t]
\centering
\includegraphics[width=\linewidth]{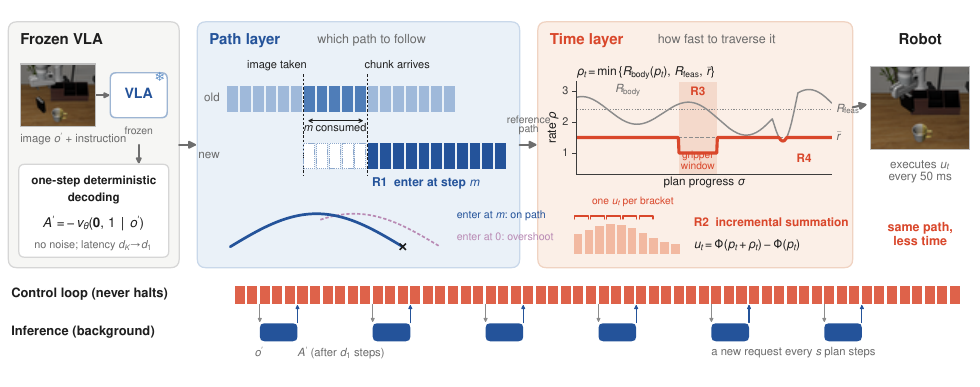}
\caption{\textbf{\pts{} overview.} The frozen VLA is queried in the background every $s$ plan steps and decoded in one deterministic step. The \emph{path layer} enters each new chunk at the progress $m$ consumed during inference (R1). The \emph{time layer} sums the planned increments (R2), runs at 1$\times$ while the gripper moves (R3), and otherwise at the largest feasible rate (R4). The control loop never halts.}
\label{fig:overview}
\end{figure}

\pts{} rests on one principle: \emph{the robot's motion is a time reparameterization of a reference path, and the reference path changes only in response to new observations.} The second clause yields the path layer (\cref{sec:path}), the first the time layer (\cref{sec:time}); \cref{sec:rate} selects the rate cap. \Cref{alg:pts} gives the complete procedure, and \cref{app:analysis} the derivations and assumptions.

\subsection{Path layer: which path to follow}
\label{sec:path}

A new chunk is a plan that starts at the moment its image $o'$ was taken. While it was being inferred, the robot consumed $m$ plan steps of the old chunk, so the part of the new plan that corresponds to ``now'' starts at its step $m$. The path should change only when new information arrives: if every switch drew a fresh random sample, the path would jump even when nothing new had been observed.
\begin{itemize}
\item \Rl{1}~\textbf{Rolling continuation.} When the result for observation $o'$ arrives, execute the plan decoded \emph{deterministically} from $o'$, entered at the progress $m$ consumed during inference. For flow-matching policies, decode with one Euler step from $x_1=0$: $A'=-v_\theta(0,1\mid o')$ \citep{chen2026simple}.
\end{itemize}

\paragraph{Why both parts of R1 matter}
Let $\Psi_A(j):=\Phi_A(E)-\Phi_A(j)$ be the displacement that chunk $A$ still commands from step $j$ to a common end $E\le H$. Let $\mathcal I:=(o',m,E)$ collect what is known at the switch. As a reference for what remains to be traveled, take $\Psi^\ast:=\Psi_{A^\ast}(m)$ for an independent policy sample $A^\ast\sim\pi(\cdot\mid o')$. A continuation rule executes a chunk $C$ entered at step $j$ and thus commands $\hat Y:=\Psi_C(j)$.

\begin{proposition}[Error decomposition]\label{prop:decomp}
If the executed pair $(C,j)$ is conditionally independent of $A^\ast$ given $\mathcal I$, then
\begin{equation}\label{eq:decomp}
\E\bigl[\|\hat Y-\Psi^\ast\|^2\mid \mathcal I\bigr]
=\underbrace{\bigl\|\E[\hat Y\mid \mathcal I]-\Psi_{\mu(o')}(m)\bigr\|^2}_{\text{bias}}
+\underbrace{\tr\Cov(\hat Y\mid \mathcal I)}_{\text{execution variance}}
+\underbrace{\tr\Cov(\Psi^\ast\mid \mathcal I)}_{\text{policy uncertainty}}.
\end{equation}
\end{proposition}
The last term does not depend on the rule, so the error is smallest when the mean plan is executed from step $m$. The two common practices each add one term. Entering at step 0 adds the bias $\|\Phi_\mu(m)-\Phi_\mu(0)\|^2$, the displacement already executed during inference, which is replayed (\cref{fig:concept}c). Executing a fresh sample entered at $m$ adds a variance equal to the policy uncertainty, doubling the error, and lets the reference path jump at a switch even for identical observations. The decomposition concerns the remaining net displacement and assumes that the $m$ steps executed during inference roughly agree with the first $m$ steps of the new plan (\cref{app:assumptions}).

\paragraph{Deterministic one-step decoding}
For the ideal velocity field $v^\ast(x,\tau\mid o)=\E[\varepsilon-A\mid x_\tau=x,o]$, the noise end gives $v^\ast(x,1\mid o)=x-\mu(o)$, so one Euler step from any $x_1$ returns the mean $\mu(o)$ \citep{chen2026simple}. A learned field only approximates this, but R1 does not need the exact mean: a deterministic output removes the variance term of \cref{eq:decomp} regardless of the approximation error. One-step decoding also shortens the latency from $d_K$ to $d_1$ control steps (\pifive{}: 3 to 2; \smol{}: 5 to 1), which relaxes the feasibility cap below.

\subsection{Time layer: how fast to traverse it}
\label{sec:time}

On a given reference path, only the timing changes. Three requirements give three rules.

\paragraph{Incremental summation (R2)}
The cumulative command should equal the cumulative path consumed so far, $\sum_{\tau<t}u^x_\tau=\Phi(p_t)-\Phi(p_0)$ for all $t$. Differencing consecutive steps shows that exactly one command rule satisfies this:
\begin{equation}\label{eq:sum}
u_t=\Phi(p_t+\rho_t)-\Phi(p_t),
\end{equation}
i.e., each command sums the planned increments it covers. Resampling the planned velocity, $u_t=f_t\,\alpha(p_t)$, instead loses $(\rho_t-f_t)\,\alpha(p_t)$ of progress per step whenever $f_t<\rho_t$ on a locally constant plan (exact expression in \cref{app:time}). Asynchronous execution enters new chunks by planned progress, so this deficit accumulates across switches. This is why asynchronous FreeSpeed fails in \cref{sec:intro}: its gain satisfies $f_t\le0.9\rho_t$.

\paragraph{Gripper synchronization (R3)}
After its command flips, the gripper moves for 2--12 control steps regardless of the arm's rate. Arm and gripper therefore keep their planned relative timing only if $\rho_t=1$ inside this window. At a constant rate $r$ over a window of $G_0$ steps starting at $\sigma_g$, the arm ends ahead of the plan by $\Phi(\sigma_g+rG_0)-\Phi(\sigma_g+G_0)$. With a planned speed of about 0.3 before closing, a 10-step window at $r=2$ leaves the end effector about 40\,mm ahead, comparable to the size of LIBERO objects. \pts{} sets $\rho_t=1$ for $1\le t-t_0\le12$ after a gripper flip at $t_0$, ending early once the fingers stop (\cref{app:identification}).

\paragraph{Rate limit (R4)}
Two constraints bound the rate. \emph{Actuator}: the end-effector displacement is proportional to the summed command (gain $\kappa=12.8$\,mm per unit) only inside a box $\mathcal U=\prod_j[-c^{j,-},c^{j,+}]$ on the translational axes. We identify the box once by open-loop pulse tests; on the Panda, $c=(2,2,3,3,3,2)$ for $(x^\pm,y^\pm,z^\pm)$ (\cref{app:identification}). The actuator cap $\Rbody(p)$ is the largest $\rho\ge1$ for which the summed command $\Phi(p+\rho')-\Phi(p)$ stays in $\mathcal U$ for all $\rho'\in[1,\rho]$. Since actions are clipped to $[-1,1]$, every plan admits $\rho=\min_{j,\pm}c^{j,\pm}=2$ on the Panda. \emph{Feasibility}: a request is issued every $s$ consumed plan steps, and its result arrives $rd$ plan steps later at rate $r$. The new chunk is entered at $m=rd$ and must last another $s+rd$ plan steps until its successor arrives. At a constant rate it therefore lasts if and only if $2rd+s\le H$, i.e., $r\le\Rfeas:=(H-s)/(2d)$ (\cref{app:time}). Outside gripper windows,
\begin{equation}\label{eq:rate}
\rho_t=\min\{\Rbody(p_t),\ \Rfeas,\ \rbar\}.
\end{equation}
Because these bounds act pointwise along the path, taking each with equality minimizes the traversal time of a fixed path. This is the separation of path and time that underlies TOPP (\cref{app:time}). As a fallback, the robot holds position if a chunk is exhausted.

\subsection{Choosing the rate cap: when does speed-up pay off?}
\label{sec:rate}

In closed loop, deviations also come from tracking lag, stale information, and the shift in the states the policy sees. All of these grow with the rate, and how much a task tolerates cannot be derived from the path. We therefore select the cap on a development set, $\rbar:=\arg\max_{r\in\{1,1.5,2,2.5,3\}}\hat\Theta_{\mathrm{dev}}(r)$. The following identity states what this choice trades off.

\begin{proposition}[Break-even success rate]\label{prop:throughput}
Compare a rate rule $r$ with 1$\times$ execution, both asynchronous with the same decoder. Let $P_r$ be the success rate and $T_s(r),T_f(r)$ the mean steps of successful and failed attempts. If $P_1>0$ and $D:=\E[T_1]+P_1(T_f(r)-T_s(r))>0$, then $\Theta(r)>\Theta(1)$ if and only if
\begin{equation}\label{eq:breakeven}
P_r>\Pstar(r):=\frac{P_1\,T_f(r)}{D}.
\end{equation}
\end{proposition}
The proof substitutes $\E[T_r]=P_rT_s(r)+(1-P_r)T_f(r)$ into $\Theta(r)>\Theta(1)$. $\Pstar$ is the success budget that speed-up buys. The faster the successful attempts, the more success may be lost; the costlier a failure (e.g., a long timeout), the tighter the budget. We call $M(r):=P_r-\Pstar(r)$ the margin. We use it to interpret the measured choice of $\rbar$ (\cref{sec:exp-rate}).

Whether deterministic decoding costs success is an empirical question. It removes the random jump at each switch and shortens the information age, but the mean can fall between modes when the action distribution is multimodal (\cref{app:path}). We therefore test one hypothesis in \cref{sec:exp-components}: \emph{under the same time-layer rules and rates, replacing $K$-step sampling by deterministic one-step decoding does not lower the success rate.}

\begin{algorithm}[t]
\caption{\pts{} (executed at every control step $t$)}
\label{alg:pts}
\small
\begin{algorithmic}[1]
\Require frozen flow-matching policy $v_\theta$; current chunk $A$ and plan position $p$; replanning interval $s$; latency $d_1$; actuator box $c$; rate cap $\rbar$
\If{an inference result $A'$ arrives ($m$ plan steps consumed since its image was taken)}
  \State $A\gets A'$, \ $p\gets m$ \Comment{R1: rolling continuation}
\EndIf
\If{$\ge s$ plan steps consumed since the last switch and no request is in flight}
  \State take image $o'$; compute $A'\gets -v_\theta(0,1\mid o')$ in the background; keep executing \Comment{R1: one-step decoding}
\EndIf
\If{gripper flipped at $t_0$, $1\le t-t_0\le 12$, and ($t-t_0\le 2$ or $|\dot q^g|>0.02$)}
  \State $\rho\gets 1$ \Comment{R3: gripper synchronization}
\Else
  \State $\rho\gets$ largest grid value in $\{1,1.25,\dots\}$, $\le\min\{\Rfeas,\rbar\}$, connected-feasible from 1 \Comment{R4: rate limit}
\EndIf
\State $u\gets\Phi(p+\rho)-\Phi(p)$; gripper takes the command at the end of the interval \Comment{R2: summation}
\If{the chunk is exhausted} \State hold position \EndIf
\State execute $u$; \ $p\gets p+\rho$
\end{algorithmic}
\end{algorithm}

\paragraph{Requirements}
\pts{} needs one open-loop identification per robot (the box $c$ and the gripper window) and one development-set selection of $\rbar$ per (policy, latency) pair. It modifies no model parameter, and for flow-matching policies it reduces the integration steps per inference from $K$ to 1.

%% file: sections/experiments.tex
\section{Experiments}
\label{sec:experiments}

We answer four questions:
\begin{itemize}[itemsep=0pt]
\item[\textbf{Q1}] How much does \pts{} raise throughput over deployable baselines, and does it cost success (\cref{sec:exp-main})?
\item[\textbf{Q2}] How much does each layer contribute, and what does each rule add (\cref{sec:exp-components})?
\item[\textbf{Q3}] Does the break-even success rate explain when speeding up pays off, across rates, actuators, and latencies (\cref{sec:exp-rate})?
\item[\textbf{Q4}] Does \pts{} still help when the target moves (\cref{sec:exp-dynamic})?
\end{itemize}

\subsection{Setup}
\label{sec:exp-setup}

\paragraph{Benchmark and policies}
We use the four LIBERO suites \citep{liu2023libero} (Spatial, Object, Goal, Long; 10 tasks each) on robosuite \citep{zhu2020robosuite} and MuJoCo \citep{todorov2012mujoco}, and LIBERO-Dynamic \citep{ren2026liberodynamic} for moving targets. We evaluate two frozen LIBERO-finetuned checkpoints released through LeRobot \citep{cadene2026lerobot}: \pifive{} \citep{pi2025pi05} and \smol{} \citep{shukor2025smolvla}, both with $H=50$. \smol{} is evaluated on Object and Goal; on Spatial its 1$\times$ success rate is about 0.5 and results are dominated by episode noise.

\paragraph{Latency}
Latency is injected in control steps, converted from measured inference times: \pifive{} $d_{10}=3$ and $d_1=2$; \smol{} $d_{10}=5$ and $d_1=1$ (\cref{app:latency}). RTC's guidance overhead is charged separately (1.79$\times$, $d=6$). During these $d$ steps the simulator advances with the executing method's commands (stop-and-wait methods hold position), and time is counted in 50\,ms control steps; \cref{sec:exp-rate} varies $d$.

\paragraph{Protocol}
The development set is used to select $\rbar$ (\pifive{}: Spatial initial states 0--14; \smol{}: Object and Goal, 0--4). The test set consists of initial states 20--29 of every task, i.e., 100 episodes per suite and method, with identical seeds across methods. The primary metric is $\Theta$ (\cref{eq:theta}); 95\% intervals come from a task-stratified paired bootstrap \citep{efron1979bootstrap} with 20{,}000 resamples. All methods use $s=8$; $\rbar=1.5$ for \pifive{} and $2$ for \smol{}, selected with \ptsk{} and reused without reselection.

\paragraph{Baselines}
\textbf{Sync-1$\times$} (stop and wait); \textbf{Async-1$\times$} (keep executing during inference; switch to the new sampled chunk at the consumed progress); \textbf{RTC} \citep{black2025rtc} (soft mask, $\beta=5$; \pifive{} only); \textbf{FreeSpeed} \citep{hu2026freespeed} (synchronous, commanded rate 2); and \textbf{Sync-2$\times$} (uniform 2$\times$ with summation, stop and wait). \textbf{Zero-latency 1$\times$} is a non-deployable reference. Two component controls isolate the layers: \textbf{\ptsk{}} (time layer with 10-step sampling) and \textbf{One-step-1$\times$} (path layer without time layer). Details are in \cref{app:baselines}.

\subsection{Q1: Throughput over deployable baselines}
\label{sec:exp-main}

\input{tables/main}

\paragraph{\pts{} has the highest throughput on every model--suite pair}
On \pifive{}, \pts{} reaches a pooled $\Theta$ of 7.96, $+1.97$ \ci{1.50}{2.44} (+33\%) over the strongest baseline, Async-1$\times$, and +71\% over Sync-1$\times$ (\cref{tab:main}); the interval against Async-1$\times$ excludes zero on every suite. On \smol{}, \pts{} reaches 6.94, $+3.54$ \ci{2.69}{4.49} (+104\%) over the strongest baseline, Sync-2$\times$. Against One-step-1$\times$, which uses the same one-step decoder \citep{chen2026simple} without the time layer, \pts{} gains $+1.72$ \ci{1.21}{2.24} (+28\%) and $+1.92$ \ci{1.03}{2.84} (+38\%) (\cref{tab:components}a). On both models \pts{} also exceeds the non-deployable zero-latency reference (\pifive{} $+1.89$ \ci{1.34}{2.44}; \smol{} $+2.49$ \ci{1.55}{3.48}), so the gain is not only from removing the waiting.

\paragraph{Speed without loss of success}
On \pifive{} the pooled success rate of \pts{} equals that of Async-1$\times$ (380/400; paired difference 0.000 \ci{-0.023}{0.023}) and is within 0.02 of One-step-1$\times$ ($-0.018$ \ci{-0.043}{0.005}). On \smol{} \pts{} has the highest success rate of all deployable methods (0.805 vs.\ at most 0.660; $+0.145$ \ci{0.085}{0.205} over Sync-2$\times$). In \cref{fig:teaser}b, the other speed-up methods (FreeSpeed, Sync-2$\times$) either trade success for speed or lose the speed gain to waiting.

\begin{figure}[!t]
\centering
\includegraphics[width=\linewidth]{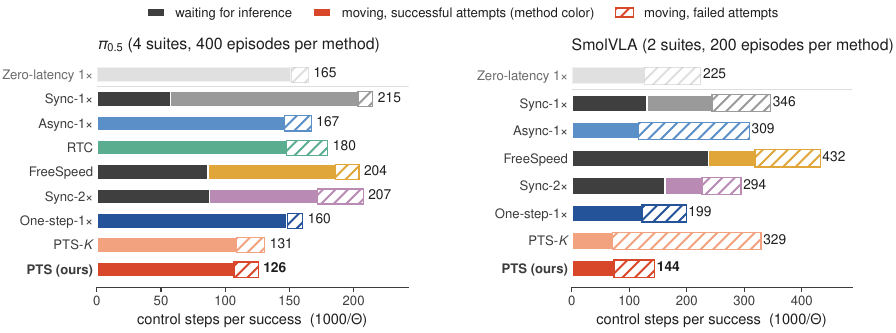}
\caption{\textbf{Where the time goes.} Control steps per success ($1000/\Theta$, lower is better), split into steps halted for inference and moving steps of successful and of failed attempts.}
\label{fig:steps}
\end{figure}

\paragraph{Where the time goes}
\Cref{fig:steps} splits each method's steps per success. Stop-and-wait methods spend 27--55\% of their steps halted, which is why speeding up synchronously (Sync-2$\times$, FreeSpeed) gains little. Asynchronous 1$\times$ execution removes the waiting but not the slow motion. \pts{} removes both while keeping the failed-attempt share small: 126 steps per success on \pifive{} versus 215 for Sync-1$\times$, and 144 versus 329 for \ptsk{} on \smol{}.

\subsection{Q2: Contribution of each layer and rule}
\label{sec:exp-components}

\input{tables/components}

\paragraph{Both layers contribute}
The \emph{time layer} raises throughput on both models (+28\% and +38\%; \cref{tab:components}a) with a similar success rate (\pifive{} 0.968$\to$0.950; \smol{} 0.790$\to$0.805). The effect of the \emph{path layer} depends on the model: it is decisive for \smol{}, whose success rate under the same timing rules rises from 0.53 with sampled decoding to 0.805 ($+0.275$ \ci{0.200}{0.350}), and positive but not significant for \pifive{} ($\Theta$ $+0.31$ \ci{-0.30}{0.92}; success $+0.005$ \ci{-0.020}{0.030}). The pooled success rates are consistent with the hypothesis of \cref{sec:rate} in all four comparisons: one-step decoding never has a lower success rate (with the time layer: \pifive{} 0.950 vs.\ 0.945, \smol{} 0.805 vs.\ 0.530; at 1$\times$: 0.968 vs.\ 0.950 and 0.790 vs.\ 0.600). Removing one-step decoding also raises the latency ($d_1\to d_K$), so these comparisons combine both effects.

\begin{figure}[!t]
\centering
\includegraphics[width=\linewidth]{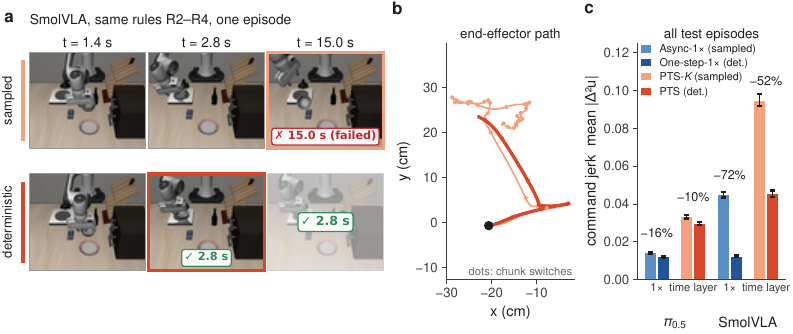}
\caption{\textbf{Deterministic decoding keeps the path from jumping.} \textbf{a},~One \smol{} LIBERO-Goal episode under the same rules R2--R4: with sampled decoding (\ptsk{}) it times out; with one-step decoding (\pts{}) it succeeds in 2.8\,s. \textbf{b},~End-effector paths (top view; dots: chunk switches). \textbf{c},~Command jerk (mean $|\Delta^2u|$, \cref{app:onestep}) over all test episodes; mean $\pm$ 95\% CI.}
\label{fig:determinism}
\end{figure}

\paragraph{Deterministic decoding keeps the path from jumping}
By \cref{prop:decomp}, sampled decoding lets the reference path jump at switches even when nothing new has been observed. \Cref{fig:determinism}c is consistent with this in the executed commands: under the same rate rule, deterministic decoding lowers command jerk by 52\% (time layer) and 72\% (1$\times$) on \smol{}, and by 10\% and 16\% on \pifive{}. The reduction tracks the success gain: large for \smol{}, small for \pifive{}. \Cref{fig:determinism}a,b shows the failure mode: after reaching the goal region, the sampled end effector wanders back and forth.

\begin{figure}[!t]
\centering
\includegraphics[width=\linewidth]{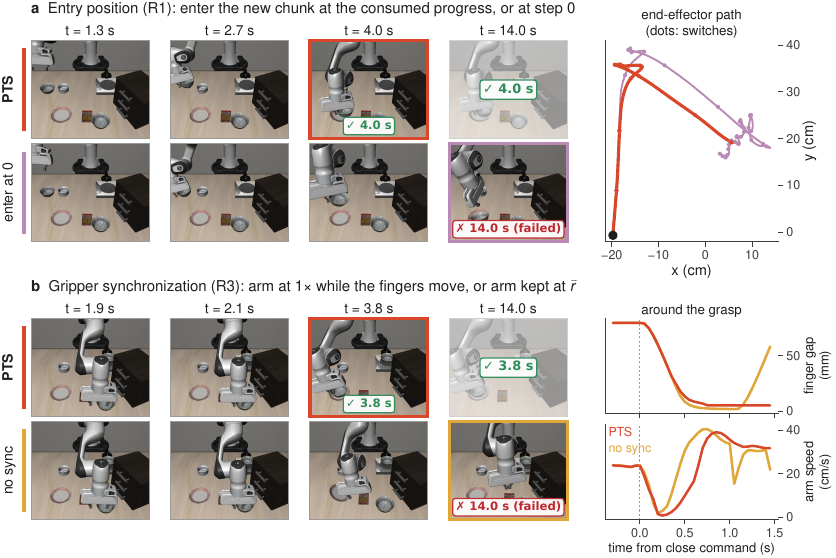}
\caption{\textbf{What the rules prevent} (\pifive{}, LIBERO-Spatial; same initial state and seed per block, columns at equal simulated times). \textbf{a},~Entering each chunk at step 0 replays motion already executed: the end effector overshoots and wanders until the time limit, whereas \pts{} goes straight to the plate. \textbf{b},~Without gripper synchronization the arm accelerates while the fingers are still closing; the grasp slips and the gripper reopens at about 1.3\,s.}
\label{fig:mechanism}
\end{figure}

\paragraph{Each replaced rule matters; entry position matters most}
Replacing the entry point of R1, R2, or R3 lowers both throughput and success (\cref{tab:components}b; one-sided bootstrap tests with Holm correction \citep{holm1979simple}). Entering new chunks at step 0 replays the plan already traversed, the bias term of \cref{prop:decomp}, and drops success to 0.71 (\cref{fig:mechanism}a). Removing gripper synchronization has a large effect on Spatial ($+3.89$ \ci{1.84}{5.93}) and a small one on Goal ($+0.93$ \ci{-0.52}{2.43}); \cref{fig:mechanism}b shows the slipping grasp it prevents. Replacing summation by resampling has the smallest effect ($+1.22$, $p_{\mathrm{Holm}}=0.027$). The same replacements on \ptsk{} lead to the same conclusions (\cref{app:extra}).

\subsection{Q3: When speeding up pays off}
\label{sec:exp-rate}

\begin{figure}[!t]
\centering
\includegraphics[width=\linewidth]{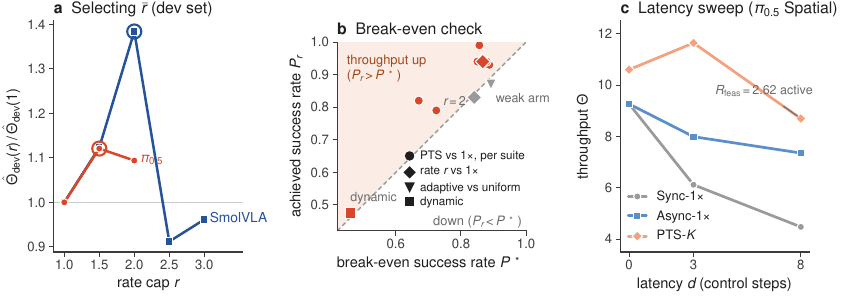}
\caption{\textbf{When speeding up pays off.} \textbf{a},~Development-set throughput relative to $r=1$ (\ptsk{}); circles: selected $\rbar$. \textbf{b},~Break-even check (\cref{prop:throughput}) on paired episodes: points above the diagonal ($P_r>\Pstar$) gain throughput (orange), points below lose it (gray). \textbf{c},~Throughput versus injected latency, \pifive{} Spatial.}
\label{fig:analysis}
\end{figure}

\paragraph{The rate cap transfers from development to test}
The development set selects $\rbar=1.5$ for \pifive{} and $\rbar=2$ for \smol{} (\cref{fig:analysis}a, \cref{tab:rate}). On the \pifive{} Spatial test set, $r=1/1.5/2$ give $\Theta=8.30/10.48/8.02$, consistent with this choice; at $r=2$ the success rate 0.83 falls below the break-even success rate $\Pstar=0.841$.

\paragraph{The break-even success rate separates gains from losses}
Comparing \pts{} with One-step-1$\times$ (same decoder, both asynchronous), all six (model, suite) pairs have $P_r>\Pstar$ and higher throughput, and the throughput gain is close to the success speed-up (\cref{tab:rate}). The only two settings with $P_r<\Pstar$ are exactly the two with lower throughput (\cref{fig:analysis}b). This explains the rate choice: \pifive{}'s 1$\times$ success rate is high and one timeout (280--520 steps) costs as much as 3--4 successes, so its budget is tight; \smol{} gains more time per success ($S=1.71$) and has a wider budget.

\paragraph{Actuator strength and latency}
With joint torque scaled by 0.7 ($\rbody=1.25$) on \pifive{} Spatial, the adaptive rate $\min\{\Rbody,\rbar\}$ does not beat uniform 1.25$\times$ ($\Theta=8.56$ vs.\ 9.18, $-0.62$ \ci{-2.22}{0.97}); its success rate (0.87) falls below the break-even success rate (0.892), so the non-uniform gain expected for a fixed path (\cref{app:time}) does not materialize. Across injected latencies, \ptsk{} has the highest throughput at every value (\cref{fig:analysis}c, \cref{tab:sweep}); the larger the latency, the more stop-and-wait execution loses. At all tested latencies $\Rfeas$ stays above $\rbar$ (2.62 at $d=8$).

\subsection{Q4: Moving targets}
\label{sec:exp-dynamic}

\begin{figure}[!t]
\centering
\includegraphics[width=0.86\linewidth]{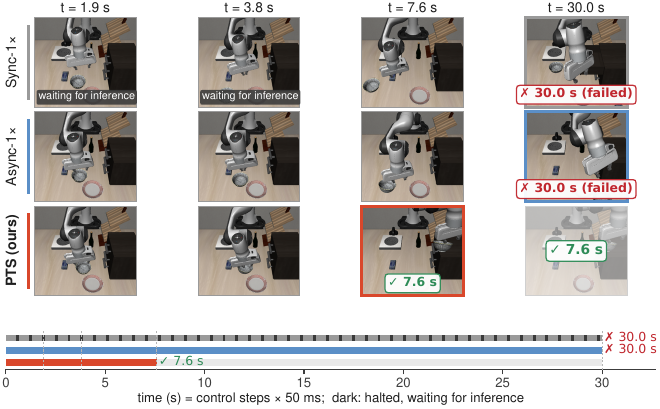}
\caption{\textbf{Moving target} (\pifive{}, LIBERO-Dynamic Goal, target at 0.06\,m/s; same seed in all rows). Neither 1$\times$ execution completes the task within the 600-step limit; \pts{} completes it in 7.6\,s. A failure case is shown in \cref{app:gallery}.}
\label{fig:dynamic}
\end{figure}

\input{tables/dynamic}

\paragraph{\pts{} improves on stop-and-wait and is comparable to asynchronous 1$\times$}
On LIBERO-Dynamic the target moves at 0.06\,m/s (10 tasks per suite, three motion levels, 30 episodes per suite and method; asynchronous methods replan every $s=2$ plan steps, stop-and-wait executes 10 steps per inference). Pooled over suites (\cref{tab:dynamic}), \pts{} improves on stop-and-wait by $+0.33$ \ci{0.04}{0.63} (+38\%), significantly on Goal and Long, and \cref{fig:dynamic} shows an episode that only \pts{} completes. Async-1$\times$, which also never halts, achieves most of this gain ($+0.28$ \ci{0.05}{0.51}), and \pts{} is comparable to it ($+0.05$ \ci{-0.23}{0.36}). On episodes both complete, \pts{} uses 21\% fewer steps, but its success rate does not improve (0.475 vs.\ 0.483) and drops on Object (0.37 vs.\ 0.50). With the policy's own success rate near 0.5, the time gained is offset by fluctuations in success (the precision-limited regime $M\approx0$ of \cref{prop:throughput}), and an $\rbar$ selected on static tasks need not suit dynamic grasping \citep{xie2026dynamicvla}.

%% file: tables/main.tex
\begin{table}[t]
\centering
\caption{\textbf{Main results.} Test-set throughput $\Theta$ (successes per 1000 control steps, waiting included; higher is better) with success rate in parentheses; 100 episodes per cell, identical initial states and seeds across methods. Bold: best deployable method per column. The zero-latency row is a non-deployable reference.}
\label{tab:main}
\small
\setlength{\tabcolsep}{4pt}
\renewcommand{\arraystretch}{1.08}
\resizebox{\linewidth}{!}{%
\begin{tabular}{@{}l ccccc c ccc@{}}
\toprule
& \multicolumn{5}{c}{\pifive{}} && \multicolumn{3}{c}{\smol{}} \\
\cmidrule(lr){2-6}\cmidrule(l){8-10}
Method & Spatial & Object & Goal & Long & \textbf{Pooled} && Object & Goal & \textbf{Pooled} \\
\midrule
\textcolor{gray}{Zero-latency 1$\times$ (ref.)} & \textcolor{gray}{9.26 (.98)} & \textcolor{gray}{7.32 (1.00)} & \textcolor{gray}{7.18 (.90)} & \textcolor{gray}{3.64 (.97)} & \textcolor{gray}{6.07 (.963)} && \textcolor{gray}{5.22 (.85)} & \textcolor{gray}{3.73 (.65)} & \textcolor{gray}{4.45 (.750)} \\
\addlinespace[2pt]
Sync-1$\times$ & 6.39 (.94) & 5.44 (1.00) & 6.13 (.94) & 2.80 (.95) & 4.65 (.958) && 2.88 (.68) & 2.90 (.60) & 2.89 (.640) \\
Async-1$\times$ & 8.30 (.94) & 7.09 (.98) & 8.59 (.97) & 3.37 (.91) & 5.99 (.950) && 2.69 (.54) & 3.88 (.66) & 3.24 (.600) \\
RTC \citep{black2025rtc} & 7.88 (.92) & 6.68 (.94) & 7.38 (.91) & 3.22 (.90) & 5.56 (.917) && -- & -- & -- \\
FreeSpeed \citep{hu2026freespeed} & 7.01 (.93) & 5.79 (.95) & 6.22 (.90) & 2.91 (.90) & 4.90 (.920) && 2.36 (.57) & 2.26 (.50) & 2.31 (.535) \\
Sync-2$\times$ & 5.53 (.80) & 6.62 (.94) & 5.92 (.82) & 3.00 (.84) & 4.82 (.850) && 4.26 (.79) & 2.61 (.53) & 3.40 (.660) \\
\midrule
\ours\textbf{\pts{} (ours)} & \textbf{10.57} (.93) & \textbf{10.09} (.99) & \textbf{10.41} (.94) & \textbf{4.67} (.94) & \textbf{7.96} (.950) && \textbf{6.82} (.82) & \textbf{7.06} (.79) & \textbf{6.94} (.805) \\
\ours\textit{\quad vs.\ best baseline} & \textit{+27\%} & \textit{+42\%} & \textit{+21\%} & \textit{+39\%} & \textit{+33\%} && \textit{+60\%} & \textit{+82\%} & \textit{+104\%} \\
\bottomrule
\end{tabular}}
\end{table}

%% file: tables/components.tex
\begin{table}[t]
\centering
\caption{\textbf{Ablations.} \textbf{(a)}~Removing a layer: pooled test throughput; in parentheses, \pts{} minus the row with 95\% paired-bootstrap interval. \textbf{(b)}~Replacing one rule at a time on \pifive{} Spatial + Goal (200 paired episodes per row); $p$: one-sided, Holm-corrected.}
\label{tab:components}
\small
\renewcommand{\arraystretch}{1.08}
\begin{minipage}[t]{\linewidth}
\centering
\begin{tabular}{@{}lcccc@{}}
\multicolumn{5}{@{}l}{\textsf{\textbf{(a) Removing a layer}}}\\
\toprule
Configuration & Path layer & Time layer & \pifive{} & \smol{} \\
\midrule
\ours\pts{} & \cmark & \cmark & \textbf{7.96} & \textbf{6.94} \\
w/o time layer (One-step-1$\times$) & \cmark & \xmark & 6.23 \;(+1.72 \ci{1.21}{2.24}) & 5.02 \;(+1.92 \ci{1.03}{2.84}) \\
w/o deterministic decoding (\ptsk{}) & \xmark & \cmark & 7.65 \;(+0.31 \ci{-0.30}{0.92}) & 3.04 \;(+3.90 \ci{2.84}{4.99}) \\
w/o both (Async-1$\times$) & \xmark & \xmark & 5.99 \;(+1.97 \ci{1.50}{2.44}) & 3.24 \;(+3.70 \ci{2.77}{4.69}) \\
\bottomrule
\end{tabular}
\end{minipage}

\vspace{7pt}
\begin{minipage}[t]{\linewidth}
\centering
\begin{tabular}{@{}lcccc@{}}
\multicolumn{5}{@{}l}{\textsf{\textbf{(b) Replacing one rule}}}\\
\toprule
Configuration & $\Theta$ & Success & \pts{} $-$ row \ci{}{} & $p_{\mathrm{Holm}}$ \\
\midrule
\ours\pts{} & \textbf{10.49} & \textbf{0.935} & -- & -- \\
R2: summation $\to$ resampling $u_t=\rho_t\alpha(p_t)$ & 9.27 & 0.905 & $+1.22$ \ci{-0.02}{2.49} & 0.027 \\
R3: gripper synchronization removed & 7.95 & 0.860 & $+2.54$ \ci{1.24}{3.83} & $<$0.001 \\
R1: entry at consumed progress $\to$ entry at step 0 & 4.28 & 0.710 & $+6.21$ \ci{4.97}{7.53} & $<$0.001 \\
\bottomrule
\end{tabular}
\end{minipage}
\end{table}

%% file: tables/dynamic.tex
\begin{table}[t]
\centering
\caption{\textbf{LIBERO-Dynamic} (target moving at 0.06\,m/s), \pifive{}, 30 episodes per suite and method: throughput $\Theta$ with success rate in parentheses.}
\label{tab:dynamic}
\small
\renewcommand{\arraystretch}{1.08}
\begin{tabular}{@{}lccccc@{}}
\toprule
Method & Spatial & Object & Goal & Long & \textbf{Pooled} \\
\midrule
Sync-1$\times$ & 1.14 (.47) & \textbf{1.28} (.57) & 1.35 (.53) & 0.06 (.03) & 0.87 (.400) \\
Async-1$\times$ & 1.34 (.53) & 1.13 (.50) & 2.60 (.73) & 0.30 (.17) & 1.15 (.483) \\
\ours\pts{} (ours) & \textbf{1.63} (.57) & 0.82 (.37) & \textbf{2.95} (.73) & \textbf{0.43} (.23) & \textbf{1.20} (.475) \\
\bottomrule
\end{tabular}
\end{table}

%% file: sections/discussion.tex
\FloatBarrier
\section{Limitations}
\label{sec:discussion}

\paragraph{Assumptions and scope of the analysis}
The time-layer rules preserve \emph{commands}; how far this carries over to poses depends on actuator linearity, which we identified in simulation (\cref{app:assumptions}). \Cref{prop:decomp} constrains only the remaining net displacement and relies on conditional independence and prefix consistency. That one-step decoding does not lower success is an empirical finding on two models, and \cref{prop:throughput} is an accounting identity that interprets rather than predicts the choice of $\rbar$.

\paragraph{Deterministic decoding}
One-step decoding is close to the conditional mean only for an accurate field, and the mean can fall between modes when a task requires choosing between clearly different actions (\cref{app:path}). $K$-step sampling with fixed noise keeps the path deterministic in that case, at the cost of latency. The measured one-step output differs from the mean of 10-step samples, which suggests that its benefit comes from being deterministic and fast rather than from averaging.

\paragraph{Rate cap and actuator}
$\rbar$ is selected on a development set and reflects the joint precision tolerance of policy, actuator, latency, and task distribution; the one-step version reuses the cap selected with \ptsk{}. On the Panda $\Rbody\ge2\ge\rbar$, so the rate is uniform outside gripper windows, and with a weakened actuator the adaptive rate did not beat a uniform one.

\paragraph{Evaluation}
All results are in simulation with injected latency; there are no real-robot experiments, where reset and timeout costs would change $\Pstar$ but not the form of \cref{prop:throughput}. The static results use one set of test initial states, and the dynamic results 30 episodes per suite, which limits statistical power. With moving targets, \pts{} does not improve on asynchronous 1$\times$ execution; selecting or adapting $\rbar$ for dynamic tasks is left open.

\section{Conclusion}
\label{sec:conclusion}

\pts{} brings the separation of path and time from classical trajectory planning to the execution of learned action chunks. The path layer lets the reference path change only with new observations, through deterministic one-step decoding and entry at the consumed progress; the time layer changes only the timing along that path, through incremental summation, gripper synchronization, and a rate limit. Without training, \pts{} raises the task throughput of frozen \pifive{} by 33\% and of \smol{} by 104\% over the strongest deployable baseline (28\% and 38\% over the same one-step decoder at 1$\times$), and exceeds zero-latency 1$\times$ execution on both. The same principle applies to any policy that outputs relative-increment chunks; testing it on real robots and with contact-aware rate caps remains future work.

%% file: appendix/analysis.tex
\section{Derivations and Assumptions}
\label{app:analysis}

This section states the assumptions behind \cref{sec:method} and gives the short derivations omitted there.

\subsection{Assumptions}
\label{app:assumptions}

\begin{itemize}
\item \textbf{Linear actuator.} For commands inside the box $\mathcal U$, the end-effector displacement equals $\kappa$ times the summed translational command, up to a bounded tracking lag. We identified this in simulation (\cref{app:identification}); it determines how far preserving \emph{commands} carries over to preserving \emph{poses}. Rotations, contacts, and gripper events are not covered.
\item \textbf{Gripper timing.} The duration of a gripper motion does not depend on the arm's rate, and the finger velocity $\dot q^g$ is observable.
\item \textbf{Prefix consistency.} The $m$ steps of the old plan executed during inference roughly agree with the first $m$ steps of the new plan, so the robot is roughly where the new plan expects it at step $m$. If the two plans point in opposite directions over these steps, entering at $m$ is only an approximation.
\end{itemize}

\subsection{Path layer}
\label{app:path}

\paragraph{Proof of \cref{prop:decomp}}
With $\mu=\mu(o')$, write $\hat Y-\Psi^\ast=(\hat Y-\E[\hat Y\mid\mathcal I])+(\E[\hat Y\mid\mathcal I]-\Psi_\mu(m))+(\Psi_\mu(m)-\Psi^\ast)$. Given $\mathcal I=(o',m,E)$, $\Psi^\ast$ is a fixed linear function of $A^\ast\sim\pi(\cdot\mid o')$, so $\E[\Psi^\ast\mid\mathcal I]=\Psi_\mu(m)$; the first and third terms have zero mean and are independent by assumption, and the second is constant. All cross terms therefore vanish in expectation, and the three squared norms give the bias, $\tr\Cov(\hat Y\mid\mathcal I)$, and $\tr\Cov(\Psi^\ast\mid\mathcal I)$. The first two vanish if and only if $\hat Y=\Psi_\mu(m)$ almost surely. Entering $\mu$ at $j$ gives bias $\|\Phi_\mu(m)-\Phi_\mu(j)\|^2$ and no variance; an independent sample entered at $m$ gives no bias and variance $\tr\Cov(\Psi^\ast\mid\mathcal I)$. Conditioning on $m$ and $E$ matters: they depend on earlier actions and gripper windows, and if they were left random, $\hat Y$ and $\Psi^\ast$ could be correlated through $m$ even when $(C,j)$ and $A^\ast$ are independent given $o'$. \qed

\paragraph{One step at the noise end}
At $\tau=1$, $x_1=\varepsilon$ is independent of $A$ given $o$, so $v^\ast(x,1\mid o)=\E[\varepsilon-A\mid x_1=x,o]=x-\mu(o)$ and $x_1-v^\ast(x_1,1\mid o)=\mu(o)$ for every $x_1$ \citep{chen2026simple}. For the learned field $v_\theta=v^\ast+e_\theta$, the output is $\mu(o)-e_\theta(0,1\mid o)$, and post-processing (unnormalization, clipping) does not commute with the expectation. Empirically, for a fixed observation and 16 ten-step samples, the distance from the one-step output to the sample mean is 2.05$\times$ (\pifive{}) and 1.17$\times$ (\smol{}) the average distance of a sample to the mean, measured on the cumulative translation of the first 10 steps. In practice the one-step output therefore differs from the sample mean, and we evaluate its effect on deployment directly (\cref{sec:exp-components}).

\paragraph{Cost of deterministic decoding}
If $\pi(\cdot\mid o)$ is multimodal, $\mu(o)$ may fall between modes, e.g., the average of two paths around an obstacle passes through it. Deterministic decoding gives up random selection between such branches. $K$-step sampling with fixed noise is also deterministic and keeps the path from jumping without taking the mean, at the cost of latency.

\subsection{Time layer}
\label{app:time}

\paragraph{Summation and resampling}
Writing the requirement $\sum_{\tau<t}u_\tau=\Phi(p_t)-\Phi(p_0)$ for $t$ and $t+1$ and subtracting gives \cref{eq:sum}; conversely, \cref{eq:sum} telescopes to the requirement. For resampling, $u^{\mathrm{sum}}_t=\rho_t\alpha(p_t)+\int_{p_t}^{p_t+\rho_t}(\alpha-\alpha(p_t))$, so
\begin{equation}
f_t\,\alpha(p_t)-u^{\mathrm{sum}}_t=(f_t-\rho_t)\,\alpha(p_t)-\int_{p_t}^{p_t+\rho_t}\bigl(\alpha-\alpha(p_t)\bigr).
\end{equation}
For a consistently moving plan and $f_t<\rho_t$, the first term opposes the planned motion at every step while the plan position still advances by $\rho_t$.

\paragraph{Gripper window}
After $i$ window steps the arm has consumed $\sum_{\tau<i}\rho_\tau$ plan steps while the gripper has moved for $i$ steps. Step-by-step synchronization, $\sum_{\tau<i}\rho_\tau=i$ for all $i$, holds if and only if $\rho\equiv1$. At a constant rate $r$, the arm reaches $\sigma_g+rG_0$ instead of $\sigma_g+G_0$, which by \cref{eq:sum} is a command offset of $\Phi(\sigma_g+rG_0)-\Phi(\sigma_g+G_0)$.

\paragraph{Actuator cap}
The set of feasible rates is in general not an interval: on one axis with $\alpha=(1,1,-1,-1)$ and $c=1.25$, $\rho=4$ is feasible ($\Phi(4)-\Phi(0)=0$) but $\rho=2$ is not. $\Rbody$ therefore uses the connected range starting at 1, and the implementation checks the grid $\{1,1.25,\dots\}$. If $|\alpha^j|\le M^{j,\pm}$, every $\rho\le\min_{j,\pm}c^{j,\pm}/M^{j,\pm}$ is feasible; with clipping, $M=1$.

\paragraph{Feasibility and information age}
A chunk is consumed for $s$ plan steps before the next request, and another $rd$ steps until that request returns; the arriving chunk is entered at $m=rd$ and must last $s+rd$ further steps, hence $2rd+s\le H$. The newest executed chunk was captured between $d$ and $2d+s/r$ control steps ago. These are continuous, constant-rate approximations; rounding of discrete requests and rate changes inside gripper windows alter the margin slightly, and the hold-position fallback covers the remainder. With one-step decoding, $d_1\le d_K$ raises $\Rfeas$ to $(H-s)/(2d_1)$ and shortens the information age.

\paragraph{Time optimality for a fixed path}
For a fixed path under pointwise bounds $r(\sigma)\le R(\sigma)$, with $R=1$ in gripper windows and $R=\min\{\Rbody,\Rfeas,\rbar\}$ elsewhere, the traversal time $\int d\sigma/r(\sigma)$ is minimized by $r=R$, uniquely almost everywhere, because the integrand decreases in $r$ and the bounds are decoupled across $\sigma$. If $\rbar\le\Rbody$ everywhere, the optimal rate is the constant $\rbar$ outside gripper windows, as on the Panda. With a weak actuator ($\rbar>\min\Rbody$) and varying planned speed, it is non-uniform: slower on fast segments, at the cap on slow ones (tested in \cref{sec:exp-rate}).

\subsection{Throughput}
\label{app:throughput}

\paragraph{Time ledger and success speed-up}
Split the control steps into waiting steps ($\rho=0$), gripper-window steps ($\rho=1$, $G$ in total), and arm steps with mean rate $\tilde r$. With consumed plan length $L=\sum_t\rho_t$,
\begin{equation}
T=|\mathcal W|+G+\frac{L-G}{\tilde r}.
\end{equation}
Stop-and-wait execution has $|\mathcal W|=d\,N_{\mathrm{replan}}$ and asynchronous execution $|\mathcal W|=0$. On episodes in which both executions succeed, with gripper fraction $g:=G/L_1$ and plan inflation $\lambda:=L_r/L_1$, the success speed-up of asynchronous execution at rate $r$ is $S(r):=\bar T_1/\bar T_r=1/\bigl(g+(\lambda-g)/r\bigr)$, which is below $1/g$ whenever $\lambda\ge1$ (about 8 on \pifive{} LIBERO-Spatial).

\paragraph{Interpretation of the break-even success rate}
When the margin $M\gg0$, the throughput gain approaches $S$ (actuator-limited regime); when $M\approx0$, the best rate lies below the actuator limit (precision-limited regime); when $M<0$, speeding up lowers throughput. $D\le0$ occurs only when failures are much shorter than successes; in our data all failures are timeouts and $D>0$. On a real robot, reset and timeout costs change $\Pstar$ but not the form of \cref{prop:throughput}.

%% file: appendix/implementation.tex
\section{Implementation Details}
\label{app:implementation}

\subsection{Actuator and gripper identification}
\label{app:identification}

The actuator box is identified with open-loop pulses: starting from rest, a constant command on one translational axis is applied for a fixed number of steps, and the ratio of actual to nominal end-effector displacement is recorded. A command level is inside the box if this ratio deviates from 1 by at most $\epsilon_{\mathrm{lin}}=0.05$. The gain is $\kappa=12.8$\,mm per action unit. \Cref{tab:box} lists the identified box for the default actuator and for the weakened actuator of \cref{sec:exp-rate} (joint-torque limits scaled by 0.7). The box is measured with single-axis pulses from rest and assumed to hold for multi-axis commands, contacts, and loads. The box is defined on the three translational axes only; rotational increments are summed by R2 in the same way as translations but are not boxed.

\begin{table}[h]
\centering
\caption{Identified actuator box $c^{j,\pm}$ (action units per step) on the Franka Panda in LIBERO.}
\label{tab:box}
\small
\begin{tabular}{@{}lcccccc@{}}
\toprule
Joint torque & $x^+$ & $x^-$ & $y^+$ & $y^-$ & $z^+$ & $z^-$ \\
\midrule
$\times1.0$ (default) & 2 & 2 & 3 & 3 & 3 & 2 \\
$\times0.7$ (weak) & 1.25 & 1.25 & 3 & 3 & 1.25 & 1.75 \\
\bottomrule
\end{tabular}
\end{table}

After a flip of the gripper command, the gripper moves for about 11 steps when empty and 2--12 steps when closing on an object. The gripper window of R3 starts at the flip step $t_0$ and lasts while $1\le t-t_0\le12$ and either $t-t_0\le2$ or the larger of the two finger joint speeds (slide joints, MuJoCo velocities) exceeds 0.02\,m/s. Inside the window the arm runs at $\rho=1$; under incremental summation the gripper takes the command at the end of the consumed interval.

\subsection{Latency model and timing}
\label{app:latency}

The first inference of an episode is not timed, identically for all methods. Afterwards, an inference requested at step $t$ returns at step $t+d$. Stop-and-wait methods hold position ($\rho=0$) during these $d$ steps, and the steps count towards $T$. Asynchronous methods keep executing. The returned chunk is computed from the observation captured at the request. Latencies are converted from measured inference times (\pifive{}: 150\,ms; \smol{}: 206\,ms for 10 steps). The one-step latency scales these times by the measured shares of the vision-language prefix and of one integration step in a full inference (\pifive{}: 0.549 and 0.051 of the full inference; \smol{}: 0.155 and 0.085), which gives $d_1=\lceil 150\cdot0.600/50\rceil=2$ and $d_1=\lceil206\cdot0.240/50\rceil=1$. RTC's guidance requires a vector--Jacobian product per integration step; we measured a 1.79$\times$ longer inference for \pifive{}, i.e., $d=6$. Other hardware gives other values of $d$; \cref{tab:sweep} varies $d$.

\subsection{Controller}
\label{app:controller}

All methods use LIBERO's default relative operational-space pose controller with one change: its input and output bounds are both multiplied by 4. This leaves the mapping from action units to commanded displacement unchanged (the same gain $\kappa$ per unit) but lets summed commands of up to 4 units pass without clipping; joint torque limits are unchanged. Each planned action is first clipped to $[-1,1]$ by the policy's post-processing and then summed by R2. The same setting is used for every method, including the 1$\times$ baselines, for which it has no effect inside $[-1,1]$, and the actuator box of \cref{tab:box} was identified under it. Checkpoints: \texttt{lerobot/pi05\_libero\_finetuned\_v044} and \texttt{HuggingFaceVLA/smolvla\_libero}. Episode limits are LIBERO's 280, 280, 300, and 520 control steps for Spatial, Object, Goal, and Long.

\subsection{Deterministic one-step decoding}
\label{app:onestep}

For both policies we call the policy's own chunk-prediction routine with the initial noise set to the zero tensor and a single integration step from $\tau=1$ to $\tau=0$, so that the output is $-v_\theta(0,1\mid o)$ followed by the policy's standard post-processing (unnormalization and clipping). Unit tests on both models verify that the output is deterministic, that exactly one denoising step is made, and that post-processing matches the original pipeline. The command jerk in \cref{fig:determinism}c is, per episode, the mean absolute second difference of the executed 6-D arm commands over moving steps.

\subsection{Baselines}
\label{app:baselines}

All methods share the policy wrapper, the latency model, the replanning interval $s=8$, the environment seeds, and the episode horizons of LIBERO.
\begin{itemize}[leftmargin=*,itemsep=2pt,topsep=2pt]
\item \textbf{Zero-latency 1$\times$}: 10-step sampling, $d=0$. A reference for what the policy achieves without any inference delay; not deployable.
\item \textbf{Sync-1$\times$}: execute $s$ plan steps, then hold for $d$ steps while the next chunk is inferred.
\item \textbf{Async-1$\times$}: request a new chunk every $s$ consumed steps and keep executing; on arrival, switch to the new sampled chunk at the consumed progress $m$ (the standard practice of skipping the stale prefix).
\item \textbf{RTC} \citep{black2025rtc}: Async-1$\times$ with guided inpainting following the paper: exponentially decaying soft mask, maximum guidance weight $\beta=5$, guidance over the entire remaining part of the previous chunk, and the frozen prefix equal to the consumed delay. The guidance overhead is charged as latency ($d=6$).
\item \textbf{FreeSpeed} \citep{hu2026freespeed}: synchronous, commanded rate $\rho=2$. Each step resamples the plan at rate $\rho$ and scales the translation and rotation increments by $f=1+(0.9\rho-1)\exp(-50\,I_{\cos})$, where $I_{\cos}$ is the directional inconsistency of adjacent translation increments. Following the paper, $f$ is computed once per cycle over the plan steps executed in that cycle. The gripper command is passed through unscaled.
\item \textbf{Sync-2$\times$}: synchronous, uniform rate 2 with incremental summation (R2), no gripper synchronization.
\item \textbf{\ptsk{}}: \pts{} with 10-step stochastic sampling instead of one-step decoding, and $d_{10}$ instead of $d_1$ in latency and $\Rfeas$.
\item \textbf{One-step-1$\times$}: Async-1$\times$ with deterministic one-step decoding and $d_1$.
\end{itemize}

\subsection{Dynamic scene}
\label{app:dynamic}

LIBERO-Dynamic \citep{ren2026liberodynamic} runs in a separate server process and latency is injected in control steps as in the static experiments; the target keeps moving during the injected inference steps, including those in which stop-and-wait methods hold position. We use target speed 0.06\,m/s, trajectory 0 of each task at the three motion levels provided by the benchmark, and 10 tasks per suite. Latency settings are as in the static experiments.

\subsection{Compute}

All test-set episodes ran on NVIDIA H100 GPUs.

%% file: appendix/results.tex
\section{Additional Results}
\label{app:extra}

\input{tables/rate}
\input{tables/sweep}

\paragraph{Per-suite differences.}
\Cref{tab:persuite} lists the paired differences of \pts{} against Async-1$\times$ (the strongest baseline for \pifive{}; for \smol{} the strongest is Sync-2$\times$ on Object and pooled, \cref{tab:main}), against \ptsk{}, against One-step-1$\times$, and against the zero-latency reference on every suite.

\begin{table}[h]
\centering
\caption{Per-suite paired differences in throughput, \pts{} minus the listed method, with 95\% task-stratified bootstrap intervals (100 episodes per suite).}
\label{tab:persuite}
\small
\setlength{\tabcolsep}{4pt}
\begin{tabular}{@{}llcccc@{}}
\toprule
Model & Suite & $-$ Async-1$\times$ & $-$ \ptsk{} & $-$ One-step-1$\times$ & $-$ Zero-latency 1$\times$ \\
\midrule
\multirow{5}{*}{\pifive{}} & Spatial & $+2.27$ \ci{0.73}{3.91} & $+0.08$ \ci{-1.70}{1.89} & $+1.41$ \ci{-0.11}{3.04} & $+1.30$ \ci{-0.44}{3.22} \\
 & Object & $+3.00$ \ci{2.51}{3.52} & $+0.06$ \ci{-0.78}{0.88} & $+3.11$ \ci{2.35}{3.77} & $+2.77$ \ci{2.10}{3.30} \\
 & Goal & $+1.82$ \ci{0.20}{3.55} & $+0.17$ \ci{-1.47}{1.80} & $+1.96$ \ci{0.36}{3.69} & $+3.23$ \ci{1.64}{4.92} \\
 & Long & $+1.30$ \ci{0.81}{1.83} & $+0.40$ \ci{-0.33}{1.12} & $+1.00$ \ci{0.45}{1.57} & $+1.03$ \ci{0.43}{1.65} \\
 & Pooled & $+1.97$ \ci{1.50}{2.44} & $+0.31$ \ci{-0.30}{0.92} & $+1.72$ \ci{1.21}{2.24} & $+1.89$ \ci{1.34}{2.44} \\
\midrule
\multirow{3}{*}{\smol{}} & Object & $+4.13$ \ci{3.00}{5.35} & $+3.49$ \ci{2.14}{4.84} & $+2.33$ \ci{1.17}{3.56} & $+1.60$ \ci{0.42}{2.84} \\
 & Goal & $+3.18$ \ci{1.67}{4.89} & $+4.30$ \ci{2.68}{6.06} & $+1.38$ \ci{0.08}{2.82} & $+3.33$ \ci{1.89}{4.95} \\
 & Pooled & $+3.70$ \ci{2.77}{4.69} & $+3.90$ \ci{2.84}{4.99} & $+1.92$ \ci{1.03}{2.84} & $+2.49$ \ci{1.55}{3.48} \\
\bottomrule
\end{tabular}
\end{table}

\paragraph{Rule replacement per suite and on \ptsk{}.}
\Cref{tab:ablation-extra} splits the full-method ablation of \cref{tab:components}b by suite and repeats the three replacements on \ptsk{} (10-step sampling). The conclusions agree: entry at step 0 is the most harmful replacement, followed by removing gripper synchronization, and then resampling.

\begin{table}[h]
\centering
\caption{Rule replacement, \pifive{}, 100 paired episodes per suite. Entries: $\Theta$ (success) and \pts{} minus row with 95\% interval; $p$ one-sided, Holm-corrected within each block.}
\label{tab:ablation-extra}
\small
\setlength{\tabcolsep}{4pt}
\begin{tabular}{@{}llccc@{}}
\toprule
Base & Configuration & $\Theta$ (success) & Base $-$ row & $p_{\mathrm{Holm}}$ \\
\midrule
\multirow{4}{*}{\pts{}, Spatial} & full & 10.57 (.930) & -- & -- \\
 & resampling & 8.61 (.880) & $+1.95$ \ci{0.19}{3.81} & 0.015 \\
 & no gripper sync & 6.67 (.790) & $+3.89$ \ci{1.84}{5.93} & $<0.001$ \\
 & entry at step 0 & 3.34 (.620) & $+7.23$ \ci{5.55}{9.03} & $<0.001$ \\
\midrule
\multirow{4}{*}{\pts{}, Goal} & full & 10.41 (.940) & -- & -- \\
 & resampling & 10.00 (.930) & $+0.41$ \ci{-1.37}{2.21} & 0.318 \\
 & no gripper sync & 9.48 (.930) & $+0.93$ \ci{-0.52}{2.43} & 0.228 \\
 & entry at step 0 & 5.47 (.800) & $+4.94$ \ci{3.09}{6.88} & $<0.001$ \\
\midrule
\multirow{4}{*}{\ptsk{}, Spatial + Goal} & full & 10.36 & -- & -- \\
 & resampling & 9.20 (.900) & $+1.16$ \ci{0.07}{2.28} & 0.019 \\
 & no gripper sync & 6.13 (.770) & $+4.23$ \ci{3.08}{5.38} & $<0.001$ \\
 & entry at step 0 & 2.15 (.465) & $+8.21$ \ci{7.17}{9.33} & $<0.001$ \\
\bottomrule
\end{tabular}
\end{table}

\paragraph{Dynamic scene per suite.}
\Cref{tab:dynamic-extra} gives the paired comparisons on LIBERO-Dynamic. ``Wins'' counts episodes in which only one of the two methods succeeds; the step ratio is computed on episodes in which both succeed (values above 1 mean the first method is faster).

\begin{table}[h]
\centering
\caption{LIBERO-Dynamic, \pifive{}, 30 paired episodes per suite (120 pooled). $\Delta\Theta$ with 95\% interval.}
\label{tab:dynamic-extra}
\small
\setlength{\tabcolsep}{4pt}
\begin{tabular}{@{}lcccc@{}}
\toprule
& Spatial & Object & Goal & Long \\
\midrule
\pts{} $-$ Sync-1$\times$ & $+0.49$ \ci{-0.42}{1.54} & $-0.46$ \ci{-1.05}{0.21} & $+1.60$ \ci{0.77}{2.74} & $+0.38$ \ci{0.18}{0.61} \\
\pts{} $-$ Async-1$\times$ & $+0.28$ \ci{-0.56}{1.28} & $-0.30$ \ci{-0.75}{0.19} & $+0.35$ \ci{-0.80}{1.69} & $+0.14$ \ci{-0.20}{0.48} \\
Async-1$\times$ $-$ Sync-1$\times$ & $+0.20$ \ci{-0.55}{0.92} & $-0.15$ \ci{-0.61}{0.34} & $+1.25$ \ci{0.57}{2.12} & $+0.24$ \ci{0.06}{0.44} \\
\bottomrule
\end{tabular}

\vspace{4pt}
\begin{tabular}{@{}lccc@{}}
\toprule
Pooled & $\Delta\Theta$ & Wins & Step ratio (both succeed) \\
\midrule
\pts{} vs.\ Sync-1$\times$ & $+0.33$ \ci{0.04}{0.63} & 25 : 16 & 1.59 (32 ep.) \\
\pts{} vs.\ Async-1$\times$ & $+0.05$ \ci{-0.23}{0.36} & 20 : 21 & 1.26 (37 ep.) \\
Async-1$\times$ vs.\ Sync-1$\times$ & $+0.28$ \ci{0.05}{0.51} & 22 : 12 & 1.04 (36 ep.) \\
\bottomrule
\end{tabular}
\end{table}

\section{Qualitative Results}
\label{app:gallery}

\paragraph{How the qualitative episodes were produced.}
All rollouts in the figures are re-executions of test-set episodes with the commands of the reported runs; a recording hook stores frames and states without changing the episode. Episodes were selected from the test logs before rendering by the outcome pattern each figure illustrates (e.g., near-median step ratio, or \pts{} succeeding where the ablated variant fails). Re-executions on a different machine can differ from the logs in episode length; for the entry-position comparison, the first candidate did not reproduce the logged failure on re-execution, and we show the next one.

\begin{figure}[p]
\centering
\includegraphics[width=\linewidth]{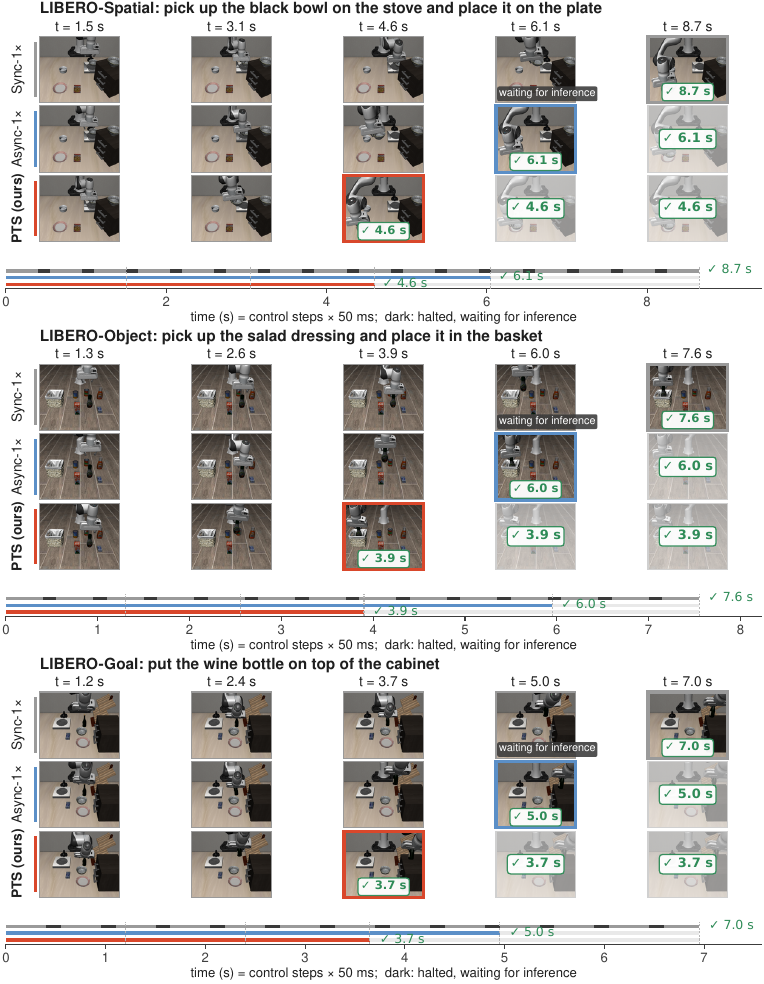}
\caption{\textbf{Static LIBERO episodes, \pifive{}.} Each block shows one task and initial state executed three ways, at the same simulated times. In every block the three executions follow visibly the same strategy; \pts{} reaches each stage earlier because it neither halts for inference nor executes at demonstration speed. Step counts: Spatial 173/121/92, Object 151/119/78, Goal 139/99/73.}
\label{fig:gallery}
\end{figure}

\begin{figure}[p]
\centering
\includegraphics[width=\linewidth]{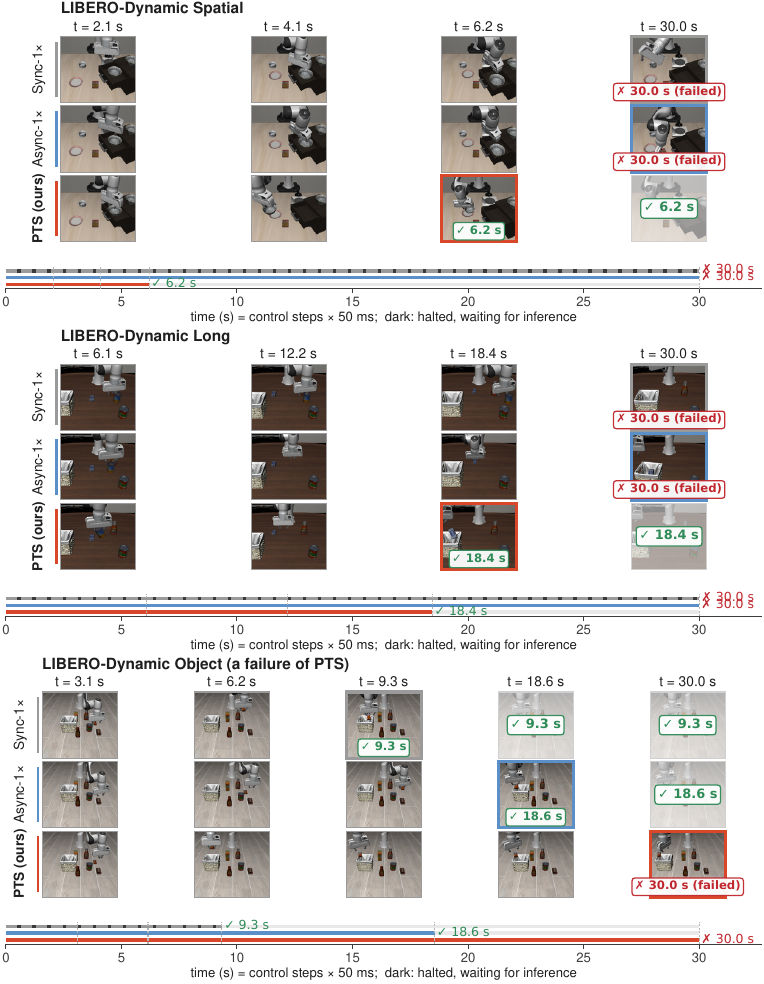}
\caption{\textbf{LIBERO-Dynamic episodes, \pifive{}.} Top and middle: episodes in which both 1$\times$ executions run out of time while \pts{} completes the task. Bottom: an Object episode in which \pts{} fails while both baselines succeed; on this suite \pts{} has the lowest success rate (\cref{tab:dynamic}).}
\label{fig:gallery-dyn}
\end{figure}

\begin{figure}[t]
\centering
\includegraphics[width=\linewidth]{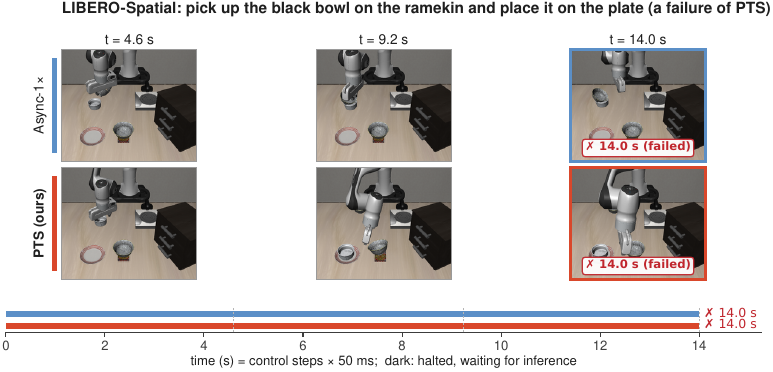}
\caption{\textbf{A static failure of \pts{}.} \pifive{} on LIBERO-Spatial; both asynchronous executions (1$\times$ and \pts{}) fail to complete the task within the 280-step limit in this episode, whereas stop-and-wait execution succeeded in the logged test run.}
\label{fig:gallery-fail}
\end{figure}

%% file: tables/rate.tex
\begin{table}[t]
\centering
\caption{Left: development-set rate selection with \ptsk{} (success rate / $\hat\Theta$; bold: selected). Right: throughput decomposition of \pts{} against One-step-1$\times$ on the test set (same decoder, both asynchronous; \cref{prop:throughput}).}
\label{tab:rate}
\small
\begin{minipage}[t]{0.56\linewidth}\centering
\setlength{\tabcolsep}{3pt}
\begin{tabular}[t]{@{}lccccc@{}}
\toprule
$r$ & 1 & 1.5 & 2 & 2.5 & 3 \\
\midrule
\multicolumn{6}{@{}l}{\pifive{} (Spatial, 150 episodes per cell)}\\
Success & 0.973 & 0.933 & 0.887 & -- & -- \\
$\hat\Theta$ & 9.27 & \textbf{10.39} & 10.14 & -- & -- \\
\midrule
\multicolumn{6}{@{}l}{\smol{} (Object + Goal, 100 episodes per cell)}\\
Success & 0.55 & 0.57 & 0.62 & 0.47 & 0.48 \\
$\hat\Theta$ & 2.84 & 3.19 & \textbf{3.93} & 2.59 & 2.73 \\
\bottomrule
\end{tabular}
\end{minipage}\hfill
\begin{minipage}[t]{0.42\linewidth}\centering
\setlength{\tabcolsep}{3pt}
\begin{tabular}[t]{@{}lcc@{}}
\toprule
Quantity & \pifive{} & \smol{} \\
\midrule
Success speed-up $S$ & 1.39 & 1.71 \\
Plan inflation $\lambda$ & 1.03 & 1.08 \\
Success $P_1\to P_r$ & .968$\to$.950 & .790$\to$.805 \\
Break-even $\Pstar$ & 0.870 & 0.699 \\
$\Theta(r)/\Theta(1)$ & 1.28 & 1.38 \\
\bottomrule
\end{tabular}
\end{minipage}
\end{table}

%% file: tables/sweep.tex
\begin{table}[h]
\centering
\caption{Latency sweep, \pifive{} Spatial, initial states 20--24, 50 episodes per cell, throughput $\Theta$ (plotted in \cref{fig:analysis}c).}
\label{tab:sweep}
\small
\begin{tabular}{@{}lccc@{}}
\toprule
$d$ & Sync-1$\times$ & Async-1$\times$ & \ptsk{} \\
\midrule
0 & 9.27 & 9.27 & \textbf{10.60} \\
3 & 6.12 & 7.99 & \textbf{11.64} \\
8 & 4.48 & 7.36 & \textbf{8.70} \\
\bottomrule
\end{tabular}
\end{table}